\documentclass[10pt,twocolumn,letterpaper]{article}
\usepackage{color}
\usepackage{iccv}
\usepackage{times}
\usepackage{epsfig}
\usepackage{graphicx}
\usepackage{amsmath}
\usepackage{amssymb}
\usepackage[T1]{fontenc}
\usepackage{booktabs}
\usepackage{bbding}
\usepackage{multirow}
\usepackage[accsupp]{axessibility}
\usepackage[table,xcdraw]{xcolor}
\usepackage{amsmath,amsthm,amssymb,amsfonts}
\usepackage{float}

\usepackage[breaklinks=true,bookmarks=false]{hyperref}

\iccvfinalcopy 

\def\iccvPaperID{****} 

\ificcvfinal\fi

\begin{document}


\title{Aperture Diffraction for Compact Snapshot Spectral Imaging}

\author{
    Tao Lv, Hao Ye, Quan Yuan, Zhan Shi, Yibo Wang, Shuming Wang\footnotemark[1], Xun Cao\thanks{Corresponding author}\\
    Nanjing University, Nanjing, China\\
    \tt\small \{lvtao, yehao, yuanquan, zhanshi, ybwang\}@smail.nju.edu.cn \\
    \tt\small \{wangshuming, caoxun\}@nju.edu.cn
}

\maketitle
\ificcvfinal\thispagestyle{empty}\fi



\begin{abstract}
    We demonstrate a compact, cost-effective snapshot spectral imaging system named Aperture Diffraction Imaging Spectrometer (ADIS), which consists only of an imaging lens with an ultra-thin orthogonal aperture mask and a mosaic filter sensor, requiring no additional physical footprint compared to common RGB cameras. Then we introduce a new optical design that each point in the object space is multiplexed to discrete encoding locations on the mosaic filter sensor by diffraction-based spatial-spectral projection engineering generated from the orthogonal mask. The orthogonal projection is uniformly accepted to obtain a weakly calibration-dependent data form to enhance modulation robustness. Meanwhile, the Cascade Shift-Shuffle Spectral Transformer (CSST) with strong perception of the diffraction degeneration is designed to solve a sparsity-constrained inverse problem, realizing the volume reconstruction from 2D measurements with Large amount of aliasing. Our system is evaluated by elaborating the imaging optical theory and reconstruction algorithm with demonstrating the experimental imaging under a single exposure. Ultimately, we achieve the  sub-super-pixel spatial resolution and high spectral resolution imaging. The code will be available at: \textcolor{red}{https://github.com/Krito-ex/CSST}.
\end{abstract}

\vspace{-0.5cm}
\section{Introduction}
\begin{figure}[t]
    \begin{center}
    \includegraphics[width=1\linewidth]{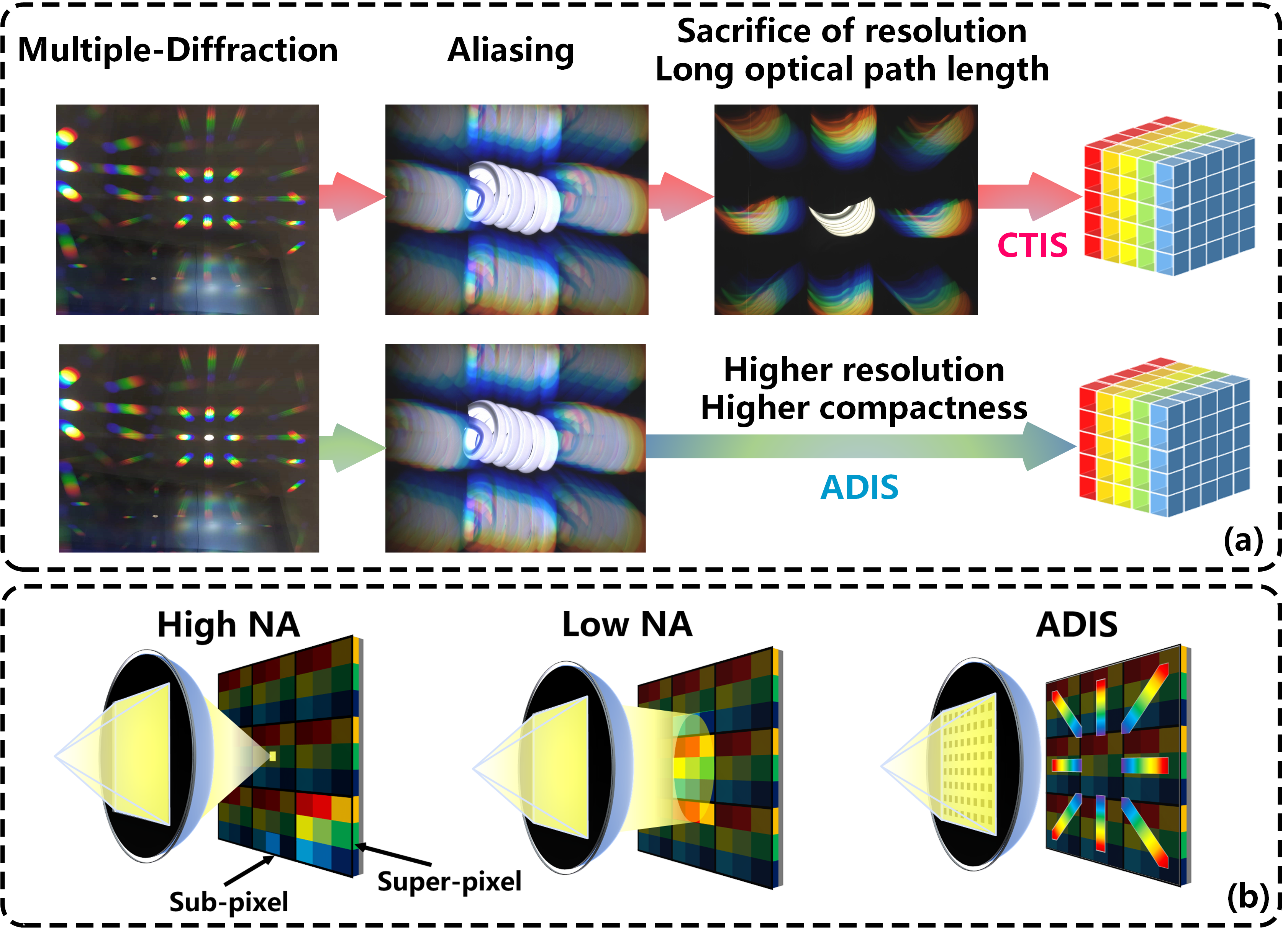}
    \end{center}
\caption{(a) illustrates the CTIS acquisition method and strategy of using long optical path with sacrificing spatial resolution, while ADIS reconstructs from aliasing; (b) depicts different imaging methods for mosaic filter sensors.}
\label{fig1}
\vspace{-0.5cm} 
\setlength{\belowcaptionskip}{-0.5cm}
\end{figure}

Snapshot spectral imaging (SSI) refers to the acquisition of a 3D spatial-spectral data cube containing spectral information at each spatial location in a single exposure~\cite{cao2021hyperspectral}. Whereas spectrum is a fundamental property that covers the physical characteristics of scenes, the visual and discriminative capabilities along the spectral and temporal dimensions will lead to unparalleled high-dimensional visual capability~\cite{yuan2023recent}. Hence, the acquisition of high temporal-spatial-spectral resolution data can provide a more comprehensive and refined observation and measurement of dynamic objects or processes.

Compared with scanning strategies of traditional imaging spectrometers along the spatial or spectral dimension, SSI methods perform specific system designs ~\cite{SDCASSI,CTIS,PMVIS,dai2014computational} based on the intrinsic sparsity of the spatial-spectral information of a scene through predefined and well-calibrated modulation or projection paradigms, which can achieve video-level capture of spectral data and have the potential for a wide range of applications in various scenarios such as combustion dynamics~\cite{hunicz2001investigation}, cellular dynamics~\cite{taruttis2015advances}, industrial monitoring~\cite{hagen2020survey}.

However, shortcomings in the compactness, the spatial-temporal-spectral resolution of the imaging system, and the robustness of the modulation limit the application of SSI  where 
portability is paramount~\cite{miniaturization, SLIM}:

SSI systems based on computational imaging methods recover the spectral cube by encoding the incident light and solving an underdetermined, sparsity-constrained inverse problem.
However, the current prevailing designs rely on bulky relay systems, prisms, or other dispersive elements that result in massive and complex optical systems~\cite{miniaturization}. 
Among these, dispersive methods exemplified by CTIS~\cite{CTIS} obviate the need for spatial modulation at the relay system's focal plane, offering the potential for compact design.
However, as shown in Figure \ref{fig1}(a), CTIS takes measures of long optical length and sacrifices spatial resolution to reduce the degree of data aliasing. 
In contrast, we propose a framework that utilizes a single mask at non-focal plane locations to achieve diffractive effects previously accomplished with complex gratings. Specific orthogonal mask diffraction can generate multiplexed spatial spectral projections to reconstruct 3D data cubes without sacrificing system integration, which consists of two sets of parallel lines with orthogonal directions. Overall, ADIS greatly improves the compactness of spectral imaging systems with the same level of integration and manufacturing cost as common RGB or monochrome cameras.

The filter array-based SSI schemes have a compact architecture, but as shown in Figure \ref{fig1}(b), the filter array itself is a sampling trade-off in spatial-spectral dimensions, sacrificing the spatial or spectral resolving ability of imaging systems~\cite{lapray2014multispectral}.
The encoding potential of the filter array, however, opens the door to an inverse solution process in ADIS. So a novel encoding scheme is adopted, treating the filter array as a sub-super-resolution encoding array with periodicity.
Further, we establish a Transformer-based deep unfolding method, CSST, with orthogonal degradation perception that simultaneously captures local contents and non-local dependencies to meet the challenge of reconstructing convincing sub-super-resolution results from highly confounded measurements.

Additionally, existing SSI technologies rely on multiple optical devices to complete optical encoding in physical space, and its accuracy in practical applications depends on the spatial-spectral mapping relationship determined by the calibration position of optical components,
while the ADIS proposed maintains spatial invariance. Under arbitrary perturbation to the aperture mask, it still uniformly maintains the mixed spectral encoding generated by the optical multiplexer to solve the movement problem faced in actual measurement.
Furthermore, when the physical parameters of the optical device are determined, the distance between the optical combiner and the sensor is the only variable that affects the spectral mapping. Therefore, ADIS reconstruction only relies on the constant parameters of the system and the distance between the system and the imaging plane, without any complicated calibration.

In summary, specific contributions are:

$\bullet$ A novel SSI framework with an optical multiplexer, enabling high-fidelity and compact snapshot spectral imaging, offering greater resilience against extraneous perturbations.

$\bullet$ A novel diffraction-projection-guided algorithm for hyperspectral reconstruction, capturing the intricate dependencies of diffraction-guided spatial-spectral mapping.

$\bullet$ A prototype device demonstrating excellent hyperspectral acquisition and reconstruction performance. 

$\bullet$ Theoretical derivation, structural analysis and necessary trade-offs for system and algorithm design.

\section{Related Work}
\vspace{-0.1cm}
\begin{figure*}[t]
\begin{center}
    \includegraphics[width=1\linewidth]{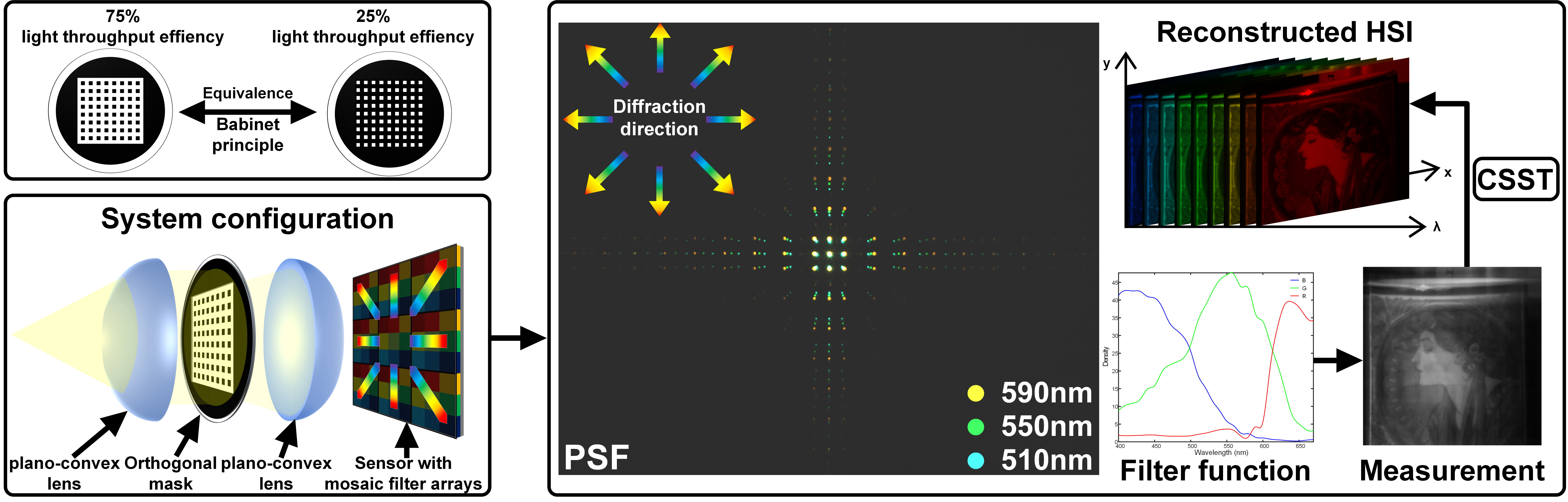}
\end{center}
\vspace{-0.3cm} 
   \caption{Illustration of ADIS architecture and reconstruction pipeline. In the upper-left, the equivalence between the two complementary masks is depicted. The PSF show in the middle is obtained by ADIS through monochromatic illumination and multiband superimposition.}
\label{fig2}
\vspace{-0.4cm} 
\setlength{\belowcaptionskip}{-0.4cm}
\end{figure*}

\textbf{Coded aperture methods} involve the utilization of a coded aperture, which selectively blocks incident light either in the focal plane or the rainbow plane~\cite{cao2016computational}. Over the past few decades, various representative systems such as CASSI~\cite{SDCASSI,DDCASSI} ,PMVIS~\cite{PMVIS} and HVIS~\cite{HVIS} have been developed to code the light field in the image plane using an occlusion mask, while employing the dispersive element to realize spectral modulation. Additionally, several improvement schemes have been proposed~\cite{DCSI,SSCSI,SCCSI}, to enhance the effectiveness of the coding process. Despite their efficacy, these systems suffer from the limitations of a bulky optical relay system and the lack of robustness in calibration due to environmental disturbances. In contrast, our system highlights the modulation robustness, which is achieved through the utilization of a clean architecture comprising solely of a mosaic sensor and lenses in combination with an optical multiplexer.

\textbf{Dispersive methods} use prisms or diffractive optics to encode the spectral information. For example, CTIS~\cite{CTIS} sacrifices spatial resolution for spectral resolution, which suffers from cone loss; or uses a single dispersion to blur the scene, but leads to a highly ill-conditioned problem and low reconstruction accuracy because the spectral encoding is only at the edges of the objects in the scene~\cite{baek2017compact}; or to further improve the compactness of the system, uses diffractive optics such as DOE to reconstruct 3D hyperspectral information based on the sparsity assumption. However, the modulation robustness of these approaches is still limited by the created anisotropic PSF~\cite{compact}. in contrast, our system preserves the system compactness with a good enhancement for potential for portable application scenarios.

\textbf{Filter-array-based methods} commonly recover desired channels by utilizing tiled spectral filter arrays in conjunction with the sensor, which incorporate a unique layout of super-pixels periodically arranged in the plane, leading to a reduction in spatial resolution with an increase in the number of sampled channels~\cite{lapray2014multispectral}. While some demosaic techniques may be used in combination with filter-array-based methods, they rely on data that is not initially captured by the sensor~\cite{mihoubi2017multispectral}. Although constrained by detector and filter dimensions, narrow-band filter-based spectrometers possess a distinct advantage in terms of miniaturization~\cite{miniaturization}. Various design solutions, such as thin-films~\cite{wang2007concept}, planar photonic crystals~\cite{pervez2010photonic}, metasurfaces~\cite{tittl2018imaging}, have been demonstrated in laboratory settings for the development of filter-array-based spectrometers. In this study, we utilize an orthogonal mask to multiplex information from a single point to different sensor locations for encoding purposes. Furthermore, our approach can be applied to other hardware solutions for mosaic encoding designs, thus extending its potential applications.

\textbf{Reconstruction Algorithm.} In the field of hyperspectral image (HSI) reconstruction, traditional iterative decoding approaches encounter significant challenges in terms of the time-consuming encoding process and the requirement for prior knowledge~\cite{yuan2016generalized,boyd2011distributed}. To address these challenges, end-to-end deep-learning approaches have been proposed and have demonstrated remarkable potential in optimizing complex ill-posed problems in various snapshot imaging systems~\cite{snapshot,TSA-net,l-net,HDnet}. Notably, $\lambda$-net~\cite{l-net} and TSA-net~\cite{TSA-net} have proposed dual-stage generative models and self-attention, respectively, to map HSI images from a single-shot measurement while modeling spatial and spectral correlation with reasonable computation cost. Recently, Transformer-based methods~\cite{MST,CST,DAUHST} have emerged as superior alternatives, outperforming previous methods and greatly improving reconstruction efficiency. Additionally, some studies have combined the strengths of both iterative and deep-learning methods by utilizing deep unfolding networks for HSI reconstruction~\cite{DNU,DAUHST}. However, most of these methods rely on a structural mathematical representation of the inverse process, which is absence in ADIS, making the above methods inapplicable or ineffective, so a Transformer-based deep unfolding method, CSST, is designed to cater the requirements of ADIS inverse solving.

\vspace{-0.3cm}
\section{System overview}
This section introduces the proposed SSI system, ADIS, covering its basic configuration, principles, and mathematical logic for determining the system imaging model and device parameters. We also discuss design trade-offs of system parameters and analyze the system's robustness to external perturbations.

\subsection{System Configuration}
Figure \ref{fig2} illustrates the configuration of our aperture diffraction imaging spectrometer system, comprising a special lens featuring an orthogonal mask on the principal plane. Alternatively, the lens can be substituted with two plano-convex lenses and orthogonal masks during experimentation. The system is completed with a mosaic array filter camera. When a field point with a smooth reflectance distribution is captured, the system disperses spectral information across different spectral bands in an orthogonal pattern. This pattern directs the information to various sub-pixel positions on the mosaic filter-encoded array. As a result, each sub-pixel on the sensor collects different bands from different spatial positions, enabling sub-super pixel resolution snapshot spectral imaging.

\subsection{Imaging Forward Model}

\begin{figure}[t]
\begin{center}
    \includegraphics[width=0.95\linewidth]{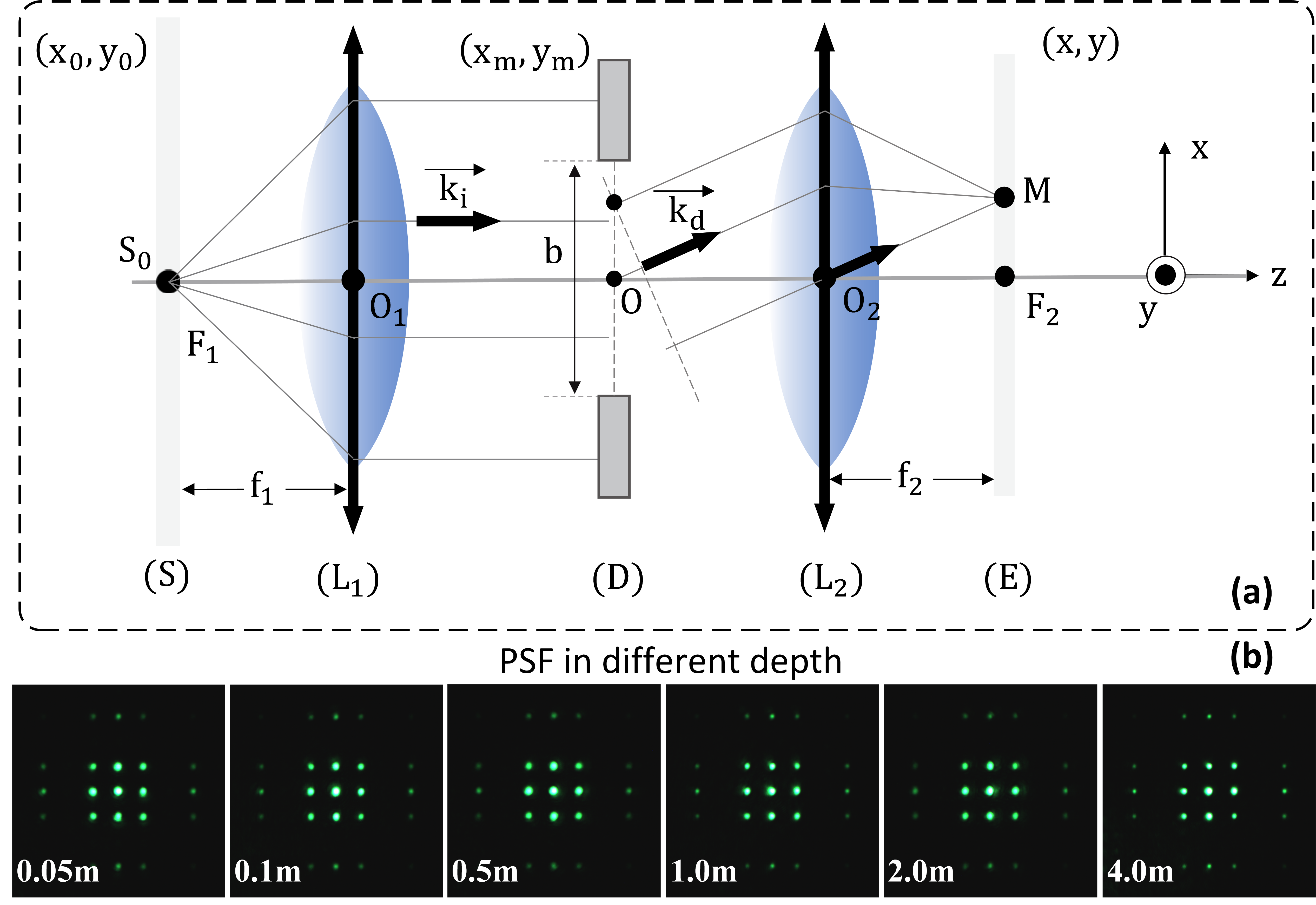}
\end{center}
\vspace{-0.4cm} 
   \caption{(a) illustrates the simplified schematic of the ADIS's profile; (b) shows the PSF of the system at different depths.}
\label{fig3}
\vspace{-0.4cm} 
\setlength{\belowcaptionskip}{-0.4cm}
\end{figure}

We now consider a multi-slit diaphragm has $N$ parallel rectangular diaphragms with rectangular square apertures of width $a$ and length $b$. The distance between two adjacent slits is $d$. A simplified schematic of ADIS is shown in Figure \ref{fig3}(a). According to the Huygens-Fresnel principle, each point on a wavefront can be considered a new secondary wave source. Thus, we can treat each rectangular square aperture as a point source for a multi-slit diaphragm. Through the amalgamation of waves generated by each of these point sources, we can effectively derive the complete wave pattern of the entire diaphragm:

\begin{equation}
\setlength{\abovedisplayskip}{0cm}
\setlength{\belowdisplayskip}{0cm}
    \label{1-1}
    {E_p} = {E_0}\frac{{\sin {\beta _1}}}{{{\beta _1}}}\frac{{\sin N{\gamma _1}}}{{\sin {\gamma _1}}}\frac{{\sin {\beta _2}}}{{{\beta _2}}}\frac{{\sin N{\gamma _2}}}{{\sin {\gamma _2}}}
\end{equation}
\noindent
Where $\theta _1$ and $\theta_2$ are the diffraction angles in x- and y-directions respectively, ${\beta _1} = \frac{1}{2}kb\sin {\theta _1}$, ${\beta _2} = \frac{1}{2}ka\sin {\theta _2}$, ${\gamma _1} = \frac{1}{2}kd\sin {\theta _1}$, ${\gamma _2} = \frac{1}{2}kd\sin {\theta _2}$. Further, by utilizing the paraxial approximation in far-field imaging, the angular relationship can be transformed into a position relationship ($\sin {\theta _1} \approx \tan {\theta _1} = \frac{{{x}}}{{{f_2}}}$, $\sin {\theta _2} \approx \tan {\theta _2} = \frac{{{y}}}{{{f_2}}}$). As a result, the intensity and position relationship of the diffraction pattern can be represented as follows:
\begin{gather}
\setlength{\abovedisplayskip}{0cm}
\setlength{\belowdisplayskip}{0cm}
\label{1-2}
   I(x,y,\lambda ) = {I_0} \cdot D(x,y,\lambda ) \cdot P(x,y,\lambda ) \\
   D(x,y,\lambda ) = \sin {c^2}(\frac{{b}}{{\lambda {f_2}}}{x})\sin {c^2}(\frac{{a}}{{\lambda {f_2}}}{y})\\
   P(x,y,\lambda ) = {\left[ {\frac{{\sin (N\frac{{\pi {d}}}{{\lambda {f_2}}}{x})}}{{\sin (\frac{{\pi {d}}}{{\lambda {f_2}}}{x})}}} \right]^2}{\left[ {\frac{{\sin (N\frac{{\pi {d}}}{{\lambda {f_2}}}{y})}}{{\sin (\frac{{\pi {d}}}{{\lambda {f_2}}}{y})}}} \right]^2}
\end{gather}
\noindent
Where the formula $D(x,y,\lambda )$ is the diffraction factor describes the diffraction effect of each rectangular square hole. $P(x,y,\lambda )$ is the interference factor describes the effect of multi-slit interference. $({x},{y})$ denotes the spatial coordinates on the receiving screen, while $f_2$ denotes the distance between the diffraction array and the sensor.

Therefore, Given our design with a lens generating orthogonal diffraction in front of the sensor, the forward model of ADIS can be considered as a combination of projection modulation and intensity encoding:
\begin{equation}
\setlength{\abovedisplayskip}{0cm}
\setlength{\belowdisplayskip}{0cm}
    \label{1-3}
    L[x,y] = \sum\limits_{\lambda  = 0}^{K - 1} {{F_\lambda }[x,y]}  \cdot Q[x,y,\lambda ]  
\end{equation}
\noindent
Where ${{F_\lambda }[x,y]}$ denotes the modulation of optical multiplexer, which is comprehensively conveyed via Equation \ref{1-2}, while $Q[x,y,\lambda ]$ denotes filtering and coding influence of mosaic filter sensors.


\begin{figure*}
\begin{center}
\includegraphics[width=1\linewidth]{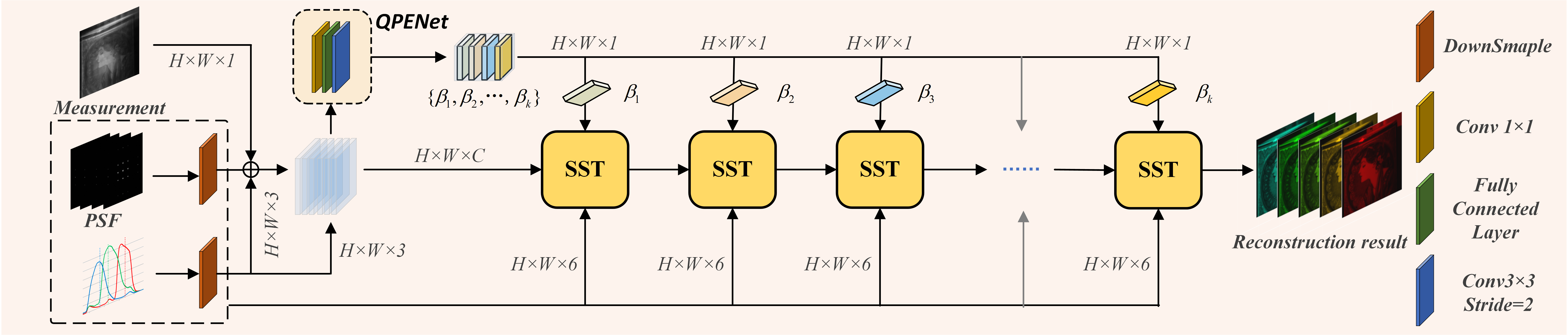}
\end{center}
\vspace{-0.4cm} 
   \caption{ Illustration of COPF architecture with k stages. Theoretically, the SST in the COPF can be replaced with a different denoiser.}
\label{fig4}
\vspace{-0.4cm} 
\setlength{\belowcaptionskip}{-0.4cm}
\end{figure*}

\subsection{Orthogonal Mask Parameters}
\vspace{-0.1cm}
Through the analytical formula of aperture diffraction in the image plane, we can analyze the relationship between different diffraction orders and the parameters of the mask. We adjust the aperture mask parameters to increase the intensity of first-order diffraction while suppressing low-order diffraction. Increasing the diffraction intensity of one order can add more spectral information to the image plane, while suppressing other diffraction orders can reduce the stray intensity information during image processing.
Whereas the dispersion function of the aperture mask is uniquely determined by the imaging focal length $f_2$ and period $d$, where the dispersion distance for the first-order diffraction is:
$\Delta {{x}_{m}}=\frac{{{f}_{2}}}{d}\left( {{\lambda }_{\max }}-{{\lambda }_{\min }} \right)$.  
Notably, expanding dispersion distance enhances system spectral resolution, yet escalating it also magnifies PSF dispersion, exacerbating reconstruction underdetermination.
Combining the effects of dispersion distance and PSF discretization on the reconstruction of the system, 
we choose appropriate square holes period $d = 10\mu m$, which is within our manufacturing capability. We calculate and compare the intensity distributions of zero-order and first-order diffraction.
 
 For the zero-order diffraction:

\begin{equation}
\setlength{\abovedisplayskip}{0cm}
\setlength{\belowdisplayskip}{0cm}
    \label{1-4}
    I={{I}_{0}}{{\left( \frac{\sin {{\beta }_{1}}}{{{\beta }_{1}}} \right)}^{2}}{{\left( \frac{\sin {{\beta }_{2}}}{{{\beta }_{2}}} \right)}^{2}}{{N}^{4}}={{I}_{0}}{{N}^{4}}
\end{equation}
 
For the first-order diffraction:

\begin{equation}
\setlength{\abovedisplayskip}{0cm}
\setlength{\belowdisplayskip}{0.3cm}
    \label{1-5}
    {I}'={{I}_{0}}{{N}^{4}}{{\left( \frac{\sin {{\beta }_{1}}}{{{\beta }_{1}}} \right)}^{2}}={{I}_{0}}{{N}^{4}}{{\left[ \frac{d}{b\pi }\sin \left( \frac{b}{d}\pi  \right) \right]}^{2}} 
\end{equation}

Let $m=\frac{d}{b}$, So ${{{I}'}}/{{{I}_{0}}}\;={{\left[ \frac{m}{\pi }\sin \left( \frac{\pi }{m} \right) \right]}^{2}}$.
According to calculations, the intensity contrast between the zero-order diffraction and the first-order diffraction depends entirely on $\frac{d}{b}$, which is the ratio between the opening aperture and the spacing of the square holes.

Furthermore, considering the diffraction pattern defined in Equation \ref{1-2}, varying m also influences diffraction patterns. 
And when we determine $a = b = 5\mu m$ for the case of $d = 10\mu m$, all the even orders will missing, which can be to reduce the projection complexity appropriately. ${I_{D_x - D_y}} = {I_0}{N^4}{A_{D_X}}{A_{D_y}}$ is expressed as the intensity relation of different diffraction levels, $D_x$, $D_y$ denote the number of diffraction orders in the orthogonal direction respectively.


\begin{equation}
\setlength{\abovedisplayskip}{0cm}
\setlength{\belowdisplayskip}{0cm}
\label{1-6}
{A_{D_x}} = {\left\{ \begin{aligned}
& 1 \quad && ,D_x = 0\\
& \frac{4}{{{D_x^2}{\pi ^2}}} \quad && ,D_x = 1,3,5,... \quad &&&     \\
& 0 \quad &&,D_x = 2,4,6,....
\end{aligned}\right.}
\end{equation}

$A_{D_x}$, $A_{D_y}$ have the same mathematical form and together define the projection form of ADIS. Furthermore, the complementary form of the $N\times N$ square aperture array can be employed using the Babinet principle, thereby elevating the light throughput efficiency from $25\%$ to $75\%$.

\subsection{Modulation Robustness}
\vspace{-0.1cm}
The maintenance of spatial invariance in optical systems is an indispensable characteristic for effectively addressing interference-related issues. While depth invariance is the main part to be considered in an imaging system, here we address the depth invariance of ADIS. Suppose a monochromatic incident wave field $u_0$ with amplitude $A_0$, phase $\phi _0$ passes through the optical multiplexer: 

\begin{equation}
\setlength{\abovedisplayskip}{0cm}
\setlength{\belowdisplayskip}{0cm}
\label{1-7}
    {u_0}({x_m},{y_m}) = {A_0}({x_m},{y_m}){e^{i{\phi _0}({x_m},{y_m})}}
\end{equation}

\noindent
An amplitude encoding and phase shift occurs by the optical multiplexer:
\begin{equation}
\setlength{\abovedisplayskip}{0.1cm}
\setlength{\belowdisplayskip}{0.1cm}
\label{1-8}
    {u_1}({x_m},{y_m}) = {u_0}({x_m},{y_m}){A_1}({x_m},{y_m}){e^{i{\phi _1}({x_m},{y_m})}}
\end{equation}

And when the ADIS is illuminated by a point light source located at depth Z. The spherical wave filed $u_0$ emitted by the source incident to the optical multiplexer can be represented by:
\begin{equation}
\setlength{\abovedisplayskip}{0cm}
\setlength{\belowdisplayskip}{0.1cm}
\label{11-9}
    {u_0}({x_m},{y_m};Z) \propto \frac{1}{\xi }{e^{ik(\xi  - Z)}}
\end{equation}
\noindent
Where $\xi  = \sqrt {{x_m}^2 + {y_m}^2 + {Z^2}} $.Since the aperture size is negligibly smaller than the imaging depth, the following relationship exists: $\xi  \approx Z$. Then the wave field $u_1$ modulated by the optical multiplexer can be expressed as:
\begin{equation}
\setlength{\abovedisplayskip}{0cm}
\setlength{\belowdisplayskip}{0cm}
\label{11-10}
    {u_1}({x_m},{y_m};Z) \propto \frac{1}{Z}{A_1}({x_m},{y_m}){e^{i\{ k(\xi  - Z) + {\phi _1}({x_m},{y_m})\} }}
\end{equation}
Since $\xi  \approx Z$, the point source is relatively close to optical infinity, and $\left( {\xi  - Z} \right)  \ll {\phi _1}({x_m},{y_m})$ holds in Equation \ref{11-10}. Then Equation \ref{11-10} can be approximated as Equation \ref{1-8}.The above derivation verifies the depth invariance of the ADIS in a specific depth range.
This derivation confirms ADIS's depth invariance within a specific range. We validated this by capturing ADIS PSFs at various depths using a $550nm$ laser (Figure \ref{fig3}(b)), revealing consistent invariance beyond the imaging focal length.

Moreover, ADIS demonstrates resilience to $(x, y)$-direction device perturbations, provided the modulation plane remains within the imaging optical path.
 Here, we assume a positional shift p in the y-direction for the Mask. Then we can get:
$ {E_p} = {E_0}\frac{{\sin {\beta _1}}}{{{\beta _1}}}\frac{{\sin N{\gamma _1}}}{{{\gamma _1}}}\frac{{\sin {\beta _2}}}{{{\beta _2}}}\frac{{\sin N{\gamma _2}}}{{{\gamma _2}}} \cdot {e^{ikp\sin \theta }}$. Taking the amplitude of the electric field can obtain $\left| {{E_p}} \right| = {E_0}\frac{{\sin {\beta _1}}}{{{\beta _1}}}\frac{{\sin N{\gamma _1}}}{{{\gamma _1}}}\frac{{\sin {\beta _2}}}{{{\beta _2}}}\frac{{\sin N{\gamma _2}}}{{{\gamma _2}}}$, which verifies the modulation robustness of the system.

\section{Hyperspectral Reconstruction}
Drawing on the benefits of self-attention for simultaneously capturing both short- and long-ranged dependencies and dynamic weighting, the Transformer architecture has demonstrated exceptional performance in a range of tasks~\cite{MST,CST,DAUHST,restormer,lSwin,shiftformer}. In parallel, the deep unfolding framework shows considerable promise through the utilization of multi-stage networks to map measurement outcomes onto the HSI, coupled with layer-by-layer optimization of the imaging system's priori model. This approach affords a more seamless integration between the depth unfolding framework and the imaging model.


In this paper, we present the Cascade Shift-Shuffle Spectral Transformer (CSST) algorithm, which is designed to improve network degradation perception by leveraging shift and shuffle operations that conform to the physical model of imaging and possess a strong perception of orthogonal diffraction projection patterns. 

\begin{figure*}
\begin{center}
\includegraphics[width=1\linewidth]{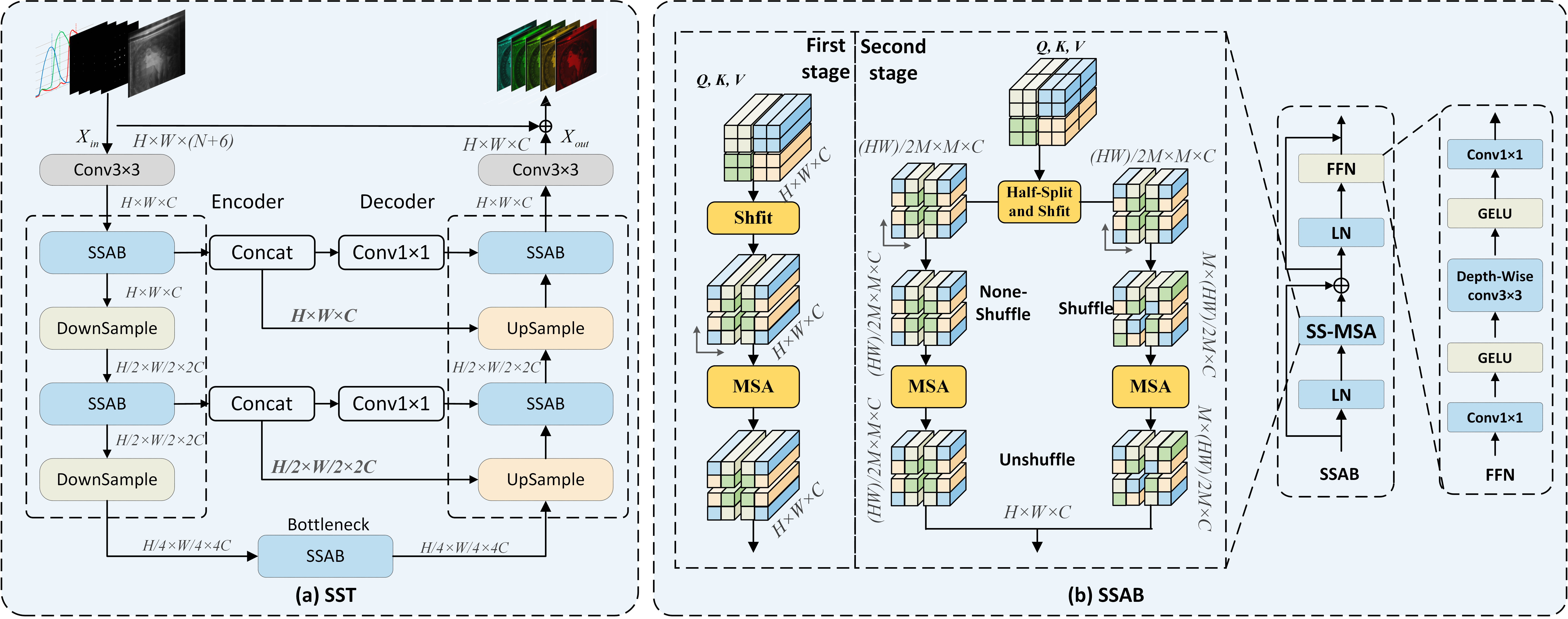}
\end{center}
\vspace{-0.4cm}
   \caption{(a) Diagram of SST with three-layer U-shaped structure; (b) SSAB consists of a FFN, a SS-SMA and two LN layers.}
\label{fig5}
\vspace{-0.4cm} 
\setlength{\belowcaptionskip}{-0.4cm}
\end{figure*}

\subsection{COPF}
To tackle the aforementioned challenges, we develop a Cascaded Orthogonal Perception Framework (COPF) that utilizes a deep unfolding framework to address the aperture diffraction degradation process. The COPF is illustrated in Figure \ref{fig4}.
First, a lightweight Quantitative Parameter Estimation network (QPENet), is designed to estimate key cues for later iterations from the system's measurements and priori information such as filter-encoded spectral response and orthogonal diffraction patterns.
Notably, the computed PSF exhibits greater spatial extent than the input filter function. To tackle data redundancy, we first downsample the PSF's spatial resolution and the filter function's channel dimension. 
Figure \ref{fig4} illustrates the architecture of QEPNet, which includes a $conv1 \times 1$, a $conv3 \times 3$, and three fully connected layers. The estimated parameter $\beta  = \{ {\beta _1},{\beta _2},...,{\beta _k}\}$ is a multichannel feature map that has the same resolution as the input features, whose number of channel layers is kept consistent with the number of iterations, allowing the estimated parameters to guide and optimize the reconstruction process pixel by pixel. Subsequently, COPF adaptively adjusts the feature map information to guide the iterative learning by inputting ${\beta}$ channel by channel at different levels of iterations. 
The initial values for the iterative process in COPF are acquired through a multi-scale integration of system measurements and prior knowledge. During the iterative learning process, the denoiser is cascaded with different cue information input directly in the iterative framework to fully utilize the guiding role of $\beta$.

\subsection{Shift-Shuffle Transformer}
The utilization of transformer models for global and local perception encounters challenges of a restricted receptive field and computationally intensive processing. So we propose a novel denoiser, Shift-Shuffle Transformer (SST) as shown in Figure \ref{fig5}, to be inserted in COPF. SST employs channel-shuffle and shift operations with fixed length (set to 1), introduced at the feature map level in the orthogonal direction. These operations improve the model's ability to perceive blending generated by aperture diffraction, while also facilitating the modeling of both short and long distances via the shift operation's function as a special token mixer. It is worth noting that the incorporation of the shift operation does not result in an increase in the total number of algorithm parameters.

\begin{figure*}
\begin{center}
\includegraphics[width=0.95\linewidth]{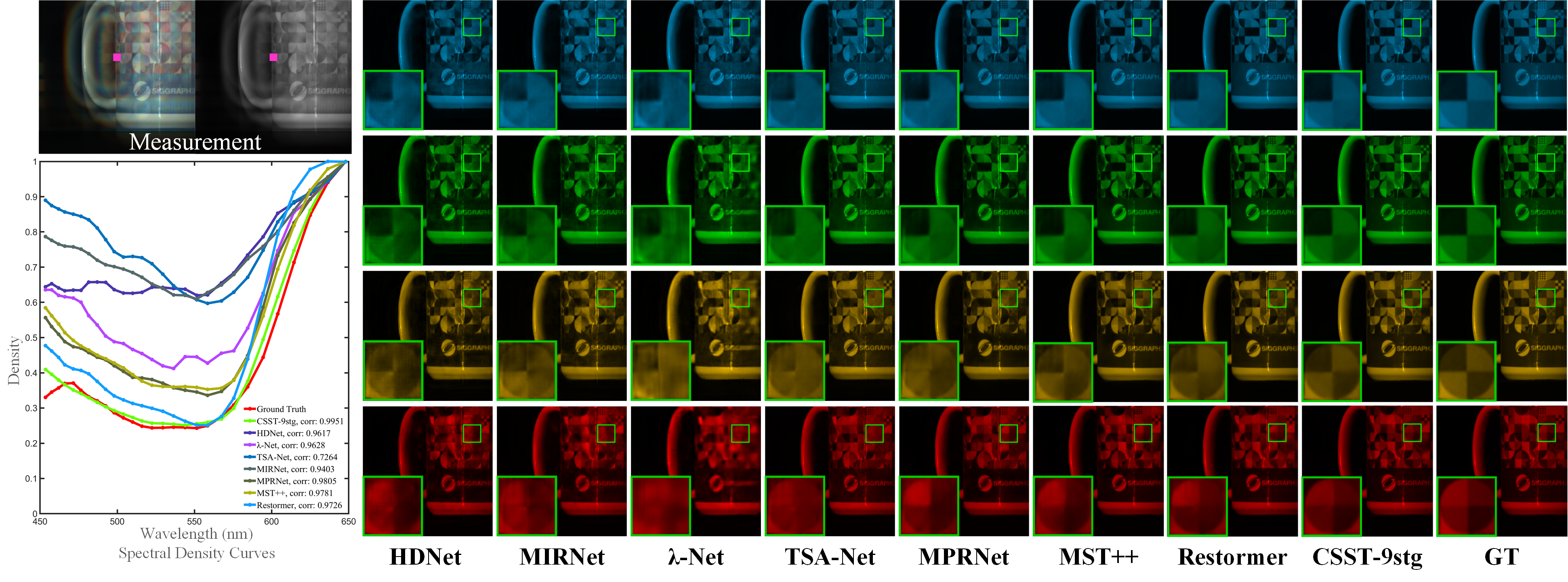}
\end{center}
\vspace{-0.4cm}
   \caption{Qualitative comparison of reconstruction results of different algorithms.  Zoomed-in patches of the HSI in the fuchsia box are presented in the lower-left of the figure. }
\label{fig6}
\vspace{-0.1cm} 
\setlength{\belowcaptionskip}{-0.2cm}
\end{figure*}

Similar to ~\cite{MST,DAUHST}, we utilize a three-layer U-shaped structure as the base framework of SST as shown in Figure \ref{fig5}(a). Firstly, SST uses a $conv3 \times 3$ to map reshaped input ${X_k}$ concatenated with stretched ${\beta _k}$, filter function $\sigma \in {\mathbb{R}^{H \times W \times 3}}$, PSF $\varsigma \in {\mathbb{R}^{H \times W \times 3}}$  into feature $X_0 \in {\mathbb{R}^{H \times W \times C}}$. Secondly, ${X_0}$ passes through the encoder, bottleneck, and decoder to be embedded into deep feature$X_f \in {\mathbb{R}^{H \times W \times C}}$. Basic unit Shift-Shuffle Attention Block (SSAB) assumes the composition of encoder and decoder.

\textbf{Shift-Shuffle Attention Block} consists of two layer normalization (LN), a SS-MSA, and a Feed-Forward Network (FFN) follows the classic design. The most important part of SSAB is Shift-Shuffle Multi-head Self-Attention(SS-MSA) with two stages: 

\textbf{First Stage}.
In the first stage of SST, only shift operations $\Upsilon \left(  \cdot  \right)$ are performed on the channels. for input tokens ${X_{in}} \in {\mathbb{R}^{H \times W \times {\rm{C}}}}$: 
\begin{equation}
\setlength{\abovedisplayskip}{0cm}
\setlength{\belowdisplayskip}{0cm}
\label{1-10}
A_1^i = {\rm softmax} (\Upsilon (\frac{{Q_1^iK_1^{{i^T}}}}{{\sqrt {{d_h}} }} + P_1^i))V_1^i
\end{equation}
Where $h=1$, $d_h=C$, $\Upsilon \left(  \cdot  \right)$ denotes shifting the input feature map by one pixel in each of its last two dimensions. And the output of first stage is $S{({X_{in}})_1} = \sum\limits_{i = 1}^h {A_1^iW_1^i}$.

\textbf{Second Stage}. Q, K, V will be split into two equal parts along the channel dimension as: ${Q_2} = [{Q_{2f}},{Q_{2s}}],{K_2} = [{K_{2f}},{K_{2s}}],{V_2} = [{V_{2f}},{V_{2s}}]$. The two parts perform different operations separately and get the corresponding results:

\begin{gather}
\setlength{\abovedisplayskip}{0cm}
\setlength{\belowdisplayskip}{0cm}
\label{1-11}
    A_{2f}^i = {\rm softmax} (\Upsilon (\frac{{Q_{2f}^iK_{2f}^{{i^T}}}}{{\sqrt {{d_h}} }} + P_{2f}^i))V_{2f}^i \\
    A_{2s}^i = {\Theta ^T}({\rm softmax} (\Upsilon (\frac{{\Theta (Q_{2s}^i)\Theta (K_{2s}^{{i^T}})}}{{\sqrt {{d_h}} }} + P_{2s}^i))\Theta (V_{2s}^i)) 
\end{gather}
Where $h=1$, $d_h=\frac{{\rm{C}}}{2}$, $\Theta \left(  \cdot  \right)$ denotes the channel shuffle operations like ShuffleNet~\cite{shufflenet} and DAHST~\cite{DAUHST}. And the output of second stage is:
\begin{equation}
\setlength{\abovedisplayskip}{0cm}
\setlength{\belowdisplayskip}{0cm}
\label{1-12}
 S{({X_{in}})_2} = \sum\limits_{i = 1}^h {A_{2f}^iW_{2f}^i}  + \sum\limits_{i = 1}^h {A_{2s}^iW_{2s}^i}   
\end{equation}

\begin{table*}[t]
\begin{center}
\resizebox{170mm}{28mm}{
\begin{tabular}{c|cccccccccccccc}
\hline
\rowcolor[HTML]{EFEFEF} 
Algorithm                                                    & Inference Time                                 & Params                                          & GFLOPS                                          & S1                                    & S2                                    & S3                                    & S4                                    & S5                                     & S6                                     & S7                                    & S8                                    & S9                                    & S10                                   & Avg                                   \\ \hline
\cellcolor[HTML]{EFEFEF}                                     &                                                &                                                 &                                                 & 29.55                                 & 27.82                                 & 24.45                                 & 31.38                                 & 27.54                                  & 27.75                                  & 24.43                                 & 31.81                                 & 33.07                                 & 24.13                                 & 28.19                                 \\
\multirow{-2}{*}{\cellcolor[HTML]{EFEFEF}HDNet~\cite{HDnet}}              & \multirow{-2}{*}{2ms}                          & \multirow{-2}{*}{2.37M}                         & \multirow{-2}{*}{144.16}                        & 0.879                                 & 0.862                                 & 0.821                                 & 0.883                                 & 0.825                                  & 0.856                                  & 0.812                                 & 0.906                                 & 0.893                                 & 0.834                                 & 0.857                                 \\ \hline
\cellcolor[HTML]{EFEFEF}                                     &                                                &                                                 &                                                 & 30.266                                & 29.09                                 & 25.10                                 & 33.04                                 & 27.52                                  & 28.46                                  & 24.66                                 & 31.94                                 & 33.31                                 & 26.22                                 & 28.96                                 \\
\multirow{-2}{*}{\cellcolor[HTML]{EFEFEF}MIRNet~\cite{MIRNet}}             & \multirow{-2}{*}{2ms}                          & \multirow{-2}{*}{2.04M}                             & \multirow{-2}{*}{14.26}                             & 0.907                                 & 0.888                                 & 0.846                                 & 0.909                                 & 0.860                                  & 0.871                                  & 0.822                                 & 0.913                                 & 0.903                                 & 0.854                                 & 0.877                                 \\ \hline
\cellcolor[HTML]{EFEFEF}                                     &                                                &                                                 &                                                 & 30.77                                 & 28.79                                 & 26.73                                 & 31.85                                 & 28.25                                  & 28.69                                  & 27.89                                 & 32.54                                 & 34.76                                 & 25.96                                 & 29.62                                 \\
\multirow{-2}{*}{\cellcolor[HTML]{EFEFEF}lambda-Net~\cite{l-net}}         & \multirow{-2}{*}{2ms}                          & \multirow{-2}{*}{32.72M}                        & \multirow{-2}{*}{23.10}                         & 0.919                                 & 0.872                                 & 0.851                                 & 0.792                                 & 0.835                                  & 0.827                                  & 0.832                                 & 0.901                                 & 0.909                                 & 0.862                                 & 0.860                                 \\ \hline
\cellcolor[HTML]{EFEFEF}                                     &                                                &                                                 &                                                 & 32.81                                 & 30.26                                 & 27.13                                 & 34.47                                 & 28.58                                  & 30.35                                  & 26.95                                 & 33.98                                 & 35.73                                 & 26.80                                 & 30.71                                 \\
\multirow{-2}{*}{\cellcolor[HTML]{EFEFEF}TSA-Net~\cite{TSA-net}}            & \multirow{-2}{*}{5ms}                          & \multirow{-2}{*}{44.23M}                        & \multirow{-2}{*}{91.19}                         & 0.948                                 & 0.923                                 & 0.900                                 & 0.923                                 & 0.901                                  & 0.913                                  & 0.865                                 & 0.941                                 & 0.926                                 & 0.914                                 & 0.915                                 \\ \hline
\cellcolor[HTML]{EFEFEF}                                     &                                                &                                                 &                                                 & 32.38                                 & 30.91                                 & 27.34                                 & 34.53                                 & 29.24                                  & 30.49                                  & 28.98                                 & 33.97                                 & 35.90                                 & 27.02                                 & 31.08                                 \\
\multirow{-2}{*}{\cellcolor[HTML]{EFEFEF}MPRNet~\cite{MPRNet}}             & \multirow{-2}{*}{3ms}                          & \multirow{-2}{*}{2.95M}                         & \multirow{-2}{*}{77.30}                         & 0.941                                 & 0.931                                 & 0.912                                 & 0.930                                 & 0.907                                  & 0.924                                  & 0.879                                 & 0.942                                 & 0.941                                 & 0.923                                 & 0.923                                 \\ \hline
\cellcolor[HTML]{EFEFEF}                                     &                                                &                                                 &                                                 & 33.75                                 & 31.78                                 & 28.87                                 & 35.51                                 & 29.95                                  & 32.34                                  & 28.01                                 & 35.03                                 & 38.53                                 & 28.49                                 & 32.23                                 \\
\multirow{-2}{*}{\cellcolor[HTML]{EFEFEF}MST++~\cite{MST++}}              & \multirow{-2}{*}{3ms}                          & \multirow{-2}{*}{1.33M}                         & \multirow{-2}{*}{17.45}                         & 0.962                                 & 0.952                                 & 0.942                                 & 0.941                                 & 0.921                                  & 0.948                                  & 0.900                                 & 0.958                                 & 0.960                                 & 0.942                                 & 0.942                                 \\ \hline
\cellcolor[HTML]{EFEFEF}                                     &                                                &                                                 &                                                 & {\color[HTML]{FE0000} \textbf{35.42}} & {\color[HTML]{329A9D} \textbf{32.62}} & {\color[HTML]{329A9D} \textbf{29.97}} & {\color[HTML]{329A9D} \textbf{36.82}} & {\color[HTML]{329A9D} \textbf{30.19}}  & {\color[HTML]{329A9D} \textbf{33.41}}  & {\color[HTML]{FE0000} \textbf{30.71}} & {\color[HTML]{329A9D} \textbf{36.00}} & {\color[HTML]{329A9D} \textbf{38.75}} & {\color[HTML]{329A9D} \textbf{28.99}} & {\color[HTML]{329A9D} \textbf{33.29}} \\
\multirow{-2}{*}{\cellcolor[HTML]{EFEFEF}Restormer~\cite{restormer}}          & \multirow{-2}{*}{10ms}                         & \multirow{-2}{*}{15.12M}                        & \multirow{-2}{*}{87.87}                         & {\color[HTML]{329A9D} \textbf{0.970}} & {\color[HTML]{329A9D} \textbf{0.959}} & {\color[HTML]{329A9D} \textbf{0.951}} & {\color[HTML]{329A9D} \textbf{0.942}} & {\color[HTML]{329A9D} \textbf{0.926}}  & {\color[HTML]{329A9D} \textbf{0.956}}  & {\color[HTML]{329A9D} \textbf{0.909}} & {\color[HTML]{329A9D} \textbf{0.961}} & {\color[HTML]{329A9D} \textbf{0.962}} & {\color[HTML]{329A9D} \textbf{0.945}} & {\color[HTML]{329A9D} \textbf{0.948}} \\ \hline
\rowcolor[HTML]{EBE5DD} 
\cellcolor[HTML]{EBE5DD}                                     & \cellcolor[HTML]{EBE5DD}                       & \cellcolor[HTML]{EBE5DD}                        & \cellcolor[HTML]{EBE5DD}                        & {\color[HTML]{329A9D} \textbf{34.72}} & {\color[HTML]{FE0000} \textbf{34.75}} & {\color[HTML]{FE0000} \textbf{31.28}} & {\color[HTML]{FE0000} \textbf{36.91}} & {\color[HTML]{FE0000} \textbf{31.601}} & {\color[HTML]{FE0000} \textbf{33.878}} & {\color[HTML]{329A9D} \textbf{30.58}} & {\color[HTML]{FE0000} \textbf{36.68}} & {\color[HTML]{FE0000} \textbf{39.29}} & {\color[HTML]{FE0000} \textbf{31.06}} & {\color[HTML]{FE0000} \textbf{34.08}} \\
\rowcolor[HTML]{EBE5DD} 
\multirow{-2}{*}{\cellcolor[HTML]{EBE5DD}\textbf{CSST-9stg (Ours)}} & \multirow{-2}{*}{\cellcolor[HTML]{EBE5DD}34ms} & \multirow{-2}{*}{\cellcolor[HTML]{EBE5DD}6.56M} & \multirow{-2}{*}{\cellcolor[HTML]{EBE5DD}70.44} & {\color[HTML]{FE0000} \textbf{0.971}} & {\color[HTML]{FE0000} \textbf{0.974}} & {\color[HTML]{FE0000} \textbf{0.964}} & {\color[HTML]{FE0000} \textbf{0.948}} & {\color[HTML]{FE0000} \textbf{0.936}}  & {\color[HTML]{FE0000} \textbf{0.964}}  & {\color[HTML]{FE0000} \textbf{0.921}} & {\color[HTML]{FE0000} \textbf{0.970}} & {\color[HTML]{FE0000} \textbf{0.969}} & {\color[HTML]{FE0000} \textbf{0.961}} & {\color[HTML]{FE0000} \textbf{0.958}} \\ \hline
\end{tabular}
}
\vspace{-0.2cm}
\end{center}
   \caption{Comparison of reconstruction results of different algorithms,Inference time, Params, FLOPS, PSNR (dB) and SSIM are reported.}
\label{tab1}
\vspace{-0.5cm} 
\setlength{\belowcaptionskip}{-0.2cm}
\end{table*}

Then we reshape the result of Equation \ref{1-12} to obtain the output ${X_{out}} \in {\mathbb{R}^{H \times W \times {\rm{C}}}}$. The global employment of shift operations, without any supplementary computational overhead, conforms with the ADIS imaging paradigm, while combined with Shuffle operations, enhances the CSST's perceptual capabilities.

\section{Experimental analysis}
Similar to ~\cite{TSA-net, yuan2021snapshot, gap-net, DGSMP, MST, CST, DAUHST}, 28 wavelengths are selected from 450nm to 650nm and derived by spectral interpolation manipulation for the HSI data. However, ADIS creates a wide-area, band-by-band form of PSF, which means that we need HSI of larger spatial size to create measurements with a certain scale of simulation to conduct experiments. Real experiments and simulation experiments with different methods and different mosaic patterns are conducted.
\subsection{simulation Experiments}
\vspace{-0.1cm}

\begin{figure*}[t]
\begin{center}
  \includegraphics[width=1\linewidth]{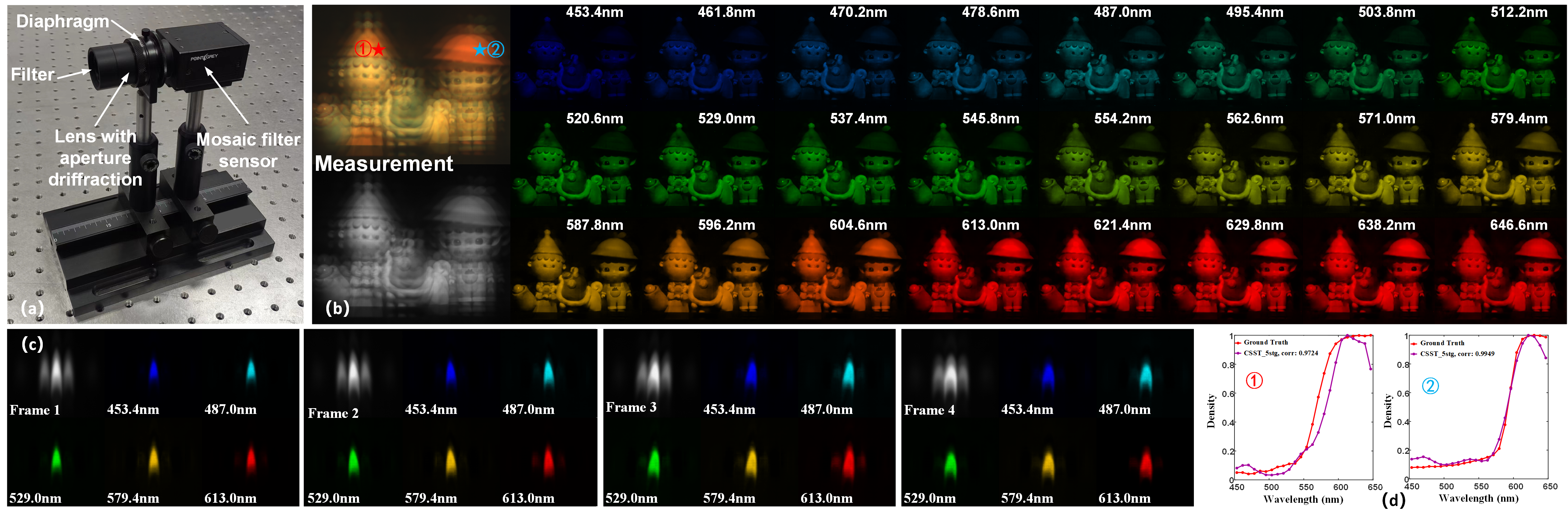}
\end{center}
\vspace{-0.4cm}
  \caption{(a) shows the prototype of ADIS; (b) illustrates the ADIS's measurements acquired from real-word and images of different spectral bands recovered by CSST-5stg; (c) shows the measurements and reconstruction results of four frames of a dynamic flame captured by ADIS; (d) Compares the recovered spectral curves and ground truth at the two markers.}
\label{fig7}
\vspace{-0.5cm} 
\setlength{\belowcaptionskip}{-0.4cm}
\end{figure*}

\textbf{Simulation Dataset.} We adopt two datasets, i.e., CAVE-1024~\cite{TSA-net} and KAIST~\cite{KAIST} for simulation experiments. The CAVE-1024 consists of 205 HSIs with spatial size 1024×1024 obtained by interpolating and splicing from the CAVE~\cite{CAVE} dataset. The KAIST dataset contains 30 HSIs of spatial size $2704\times3376$. 10 scenes from the KAIST dataset are selected for testing, while the CAVE-1024 dataset and another 20 scenes from the KAIST dataset are selected for training.

\textbf{Implementation Details.} The dispersion step of the primary diffraction is $0.5$ spatial pixels, while the simulation experiment is deployed in the range of $400nm$ to $670nm$, which means that $586 \times586 \times28$ data cubes are needed to generate $256 \times256$ resolution measurements for conducting experiments while preserving the tertiary diffraction.
We implement CSST by Pytorch. All CSST models are trained with Adam~\cite{ADAM} optimizer (${\beta _1} = 0.9$ and ${\beta _2} = 0.999$) using Cosine Annealing scheme~\cite{sgdr} for 300 epochs on an RTX 3090 GPU. The initial learning rate is $4 \times {10^{ - 4}}$.

\textbf{Quantitative Analysis.} Table \ref{tab1} compares the results of CSST and 7 methods including four reconstruction methods(lambda-Net~\cite{l-net}, TSA-Net~\cite{TSA-net}, HDNet~\cite{HDnet} and MST++~\cite{MST++}), three Super-resolution algorithms (Restormer~\cite{restormer}, MPRNet~\cite{MPRNet}, MIRNet\cite{MIRNet}) on 10 simulation scenes. CSST shows the best experimental results on the ADIS spectral reconstruction task, i.e., 34.08dB in PSNR and 0.958 in SSIM. CSST-9stg significantly outperforms two recent SOTA methods Restormer and MST++ by 0.79dB and 1.85dB, demonstrating the effectiveness and acceptability of the imaging system.

\textbf{Qualitative Analysis.} Figure \ref{fig6} illustrates the comparative performance of our CSST and other methods in the HSI reconstruction of ADIS on the same scene. Visual inspection of the image reveals that the CSST-9stg method provides more intricate details, sharper textures, and well-defined structures. Conversely, the previous approaches produce either overly smooth results that compromise the underlying structure or introduce color artifacts and speckled textures. Moreover, the lower left corner of the figure presents the spectral profile of the intensity-wavelength corresponding to the fuchsia square. The CSST-9stg spectral profile exhibits the highest correlation and overlap with the reference curve, demonstrating the superiority of our approach in achieving spectral dimensional consistency reconstruction and the effectiveness of ADIS.

\subsection{Real HSI Reconstruction}

\textbf{Implementation Details.} Firstly, we develop a prototype system utilizing an orthogonal mask with $25\% $ light-throughput and a Bayer array, as illustrated in Figure 6 (top left). This prototype includes additional filters with a wavelength range of $450nm-650nm$ to restrict the operating band, and an adjustable diaphragm. The small footprint of the system enables high-dimensional information acquisition. The orthogonal mask utilized in prototype is created by overlapping two sets of parallel lines, each with a width and interval of 5µm, and the width uniformity accuracy is 0.2µm. The mask has daiameter of $25.4mm$ and includes a $12mm\times 12mm$ modulation surface. It is custom-priced at \$80 per unit, with costs below \$5 per unit for commercial volume production. Once the physical setup of the system was determined, all projection relationships by could be easily computed by Equation \ref{1-2}  even under the disturbances.


\textbf{Training Dataset.} We trian CSST-5stg with the real configuration on CAVE-1024 and KAIST datasets jointly. Meanwhile, to address the disparity between real-world experiments and simulations arising from inherent noise and our omission of higher-order low-intensity diffraction, we incorporated randomized noise into the training data for model training, thereby bridging the aforementioned gap.

\textbf{Experimental analysis.} 
The performance of real HSI reconstruction is demonstrated in Figure \ref{fig7}(b), which presents the measurements of a spatial size of $1056\times 1536$ captured from real-world scenes, and the corresponding recovered spectral data of a spatial size of $1056\times 1536\times 24$. The reconstructed spectral data exhibit well-structured content, clear textures, and minimal artifacts. Notably, the predicted spectral curves of the two marker points closely match the curves collected using a point spectrometer. These results provide compelling evidence for the correctness and effectiveness of the mathematical model, design framework, and reconstruction algorithm architecture.

\textbf{Dynamic Performance Verification.} 
In Figure \ref{fig7}(c), the snapshot performance of ADIS is demonstrated through dynamic flame video reconstruction(35 fps).

\subsection{Ablation Study}

Here we further conduct ablation experiments on each effective component of the CSST algorithm proposed in this paper to demonstrate the necessity of the components used in the algorithm.

\vspace{-0.2cm} 
\begin{table}[h]
\centering
\resizebox{0.60\columnwidth}{!}{
\begin{tabular}{cccc}
\hline
\rowcolor[HTML]{EFEFEF} 
COPF & Shift (x,y) & PSNR(dB) & SSIM \\ \hline
\Checkmark  & \XSolidBrush          &    27.77      &    0.870  \\
\Checkmark  & (1,1)                 &    \textbf{28.74}   &    \textbf{0.885}  \\
\XSolidBrush   & (1,1)              &      27.83    &   0.871   \\
\Checkmark  & (2,2)                 &    28.44      &    0.879  \\
\Checkmark  & (3,3)                 &    28.19      &    0.873  \\
\Checkmark  & (4,4)                 &    28.30      &    0.876  \\
\Checkmark  & (5,5)                 &    28.01      &    0.868  \\ \hline
\end{tabular}}
\vspace{+0.2cm} 
\caption{Break-down ablation results in SS-MSA and COPF, Performance comparison of CSST with different shift step.}
\label{tab2}
\vspace{-0.3cm} 
\setlength{\belowcaptionskip}{-0.3cm}
\end{table}


We first remove the global shift operations in SS-MSA and COPF from CSST-3stg to conduct the break-down ablation as shown in Table \ref{tab2}.
Then We further conduct a comparative analysis to investigate the impact of the shift step size utilized in the shift operations on the effectiveness of CSST reconstruction. The results presented in Table \ref{tab2} demonstrate a decreasing trend in the reconstruction efficacy of CSST with increasing shift step size. However, it is noteworthy that all the CSST algorithms with the shift operation outperform the algorithm that lacks the shift operations.

\subsection{Simulation with Different Mosaic patterns}

\begin{figure}[h]
\begin{center}
    \includegraphics[width=1\linewidth]{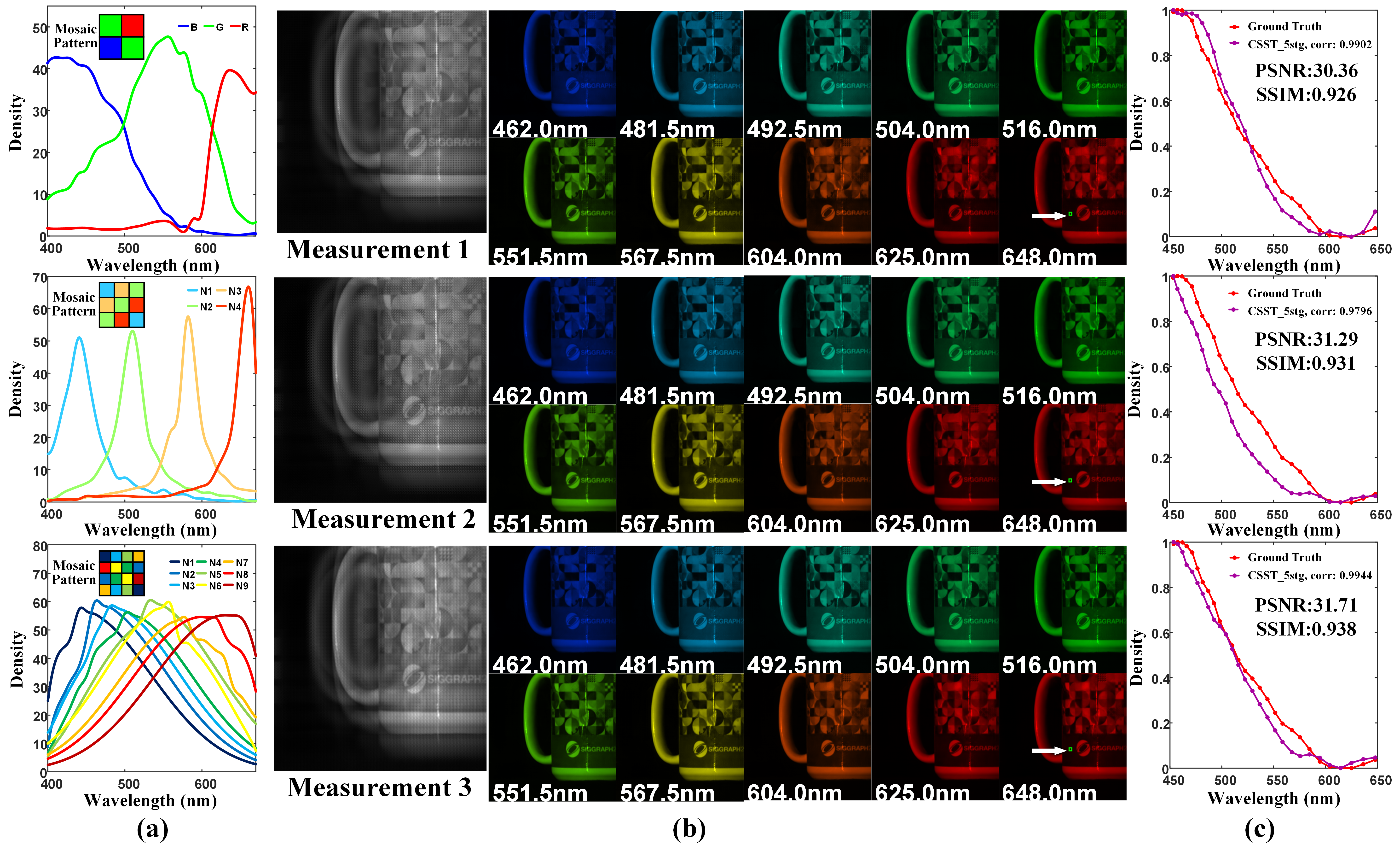}
\end{center}
\vspace{-0.4cm}
  \caption{(a) Different mosaic patterns with different filter functions; (b) illustrates the reconstruction results of ADIS combined with different mosaic filter sensor simulations; (c)  illustrates recovered spectral curves and ground-truth in the green box.}
\label{fig8}
\vspace{-0.4cm} 
\setlength{\belowcaptionskip}{-0.4cm}
\end{figure}

The current section aims to investigate the adaptability of the ADIS architecture with diverse mosaic arrays, and here we utilize CSST-5stg for comparative experiments. The experimental setups employed in this study remain consistent with Section 5.1, with the exception of the encoding form of the sensor mosaic array, which is altered. Three distinct mosaics, including a $2\times 2$ pattern with 3 channels, a $3\times 3$ pattern with 4 channels, and a $4\times 4$ pattern with 9 channels, are utilized for comparative experimentation, as demonstrated in Figure \ref{fig8}. With the improvement of filter encoding capability, the imaging performance of ADIS can be further improved.

\subsection{Evaluation of Real System}
\begin{figure}[t]
\begin{center}
  \includegraphics[width=1\linewidth]{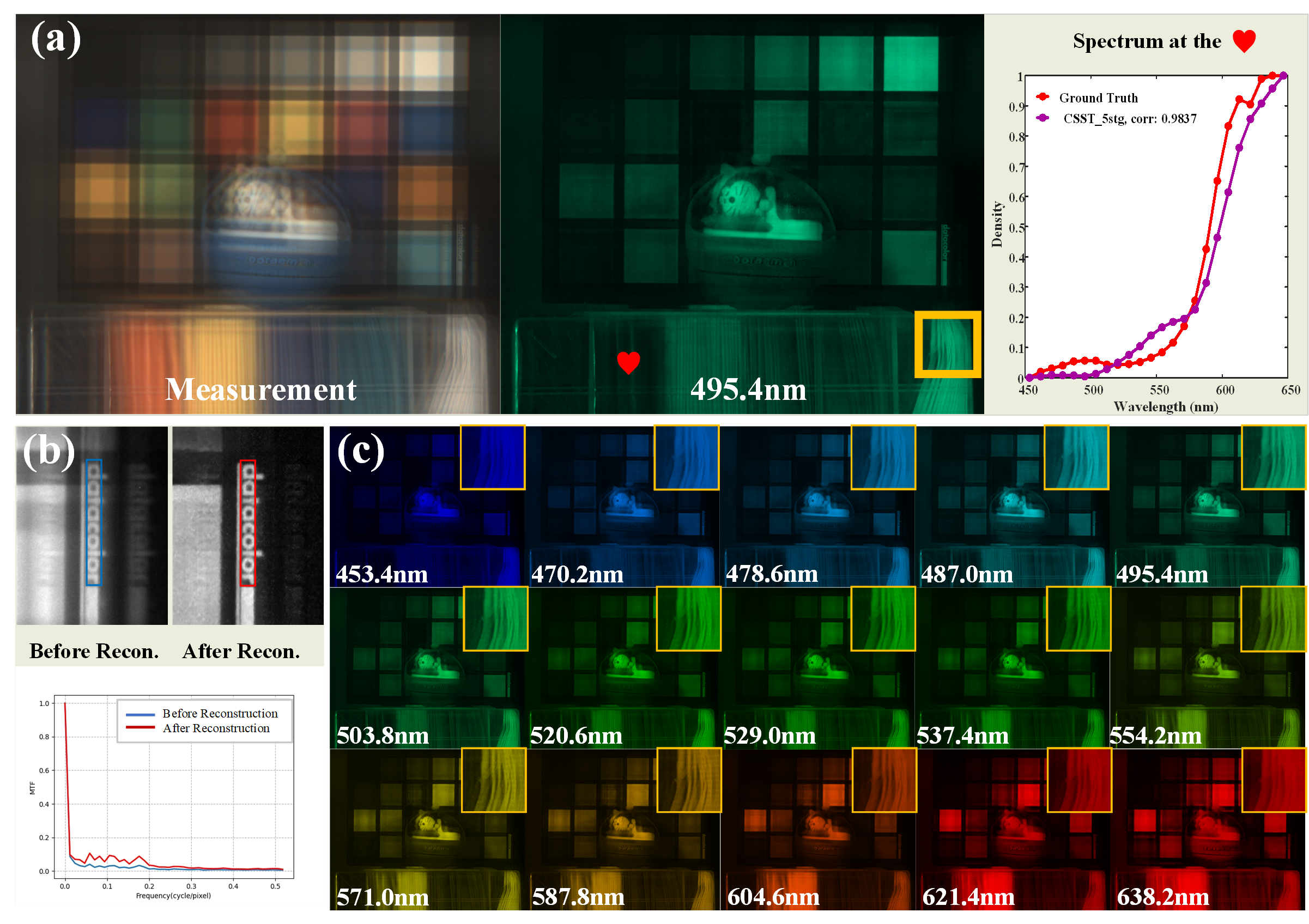}
\end{center}
\vspace{-0.3cm}
  \caption{(a) Measurement and reconstruction results of a spectral-interesting, texture-complex scene, with a comparison of reconstructed spectra and ground truth spectra at the heart-shaped markers; (b) MTF comparison of the images before and after reconstruction; (c) reconstruction results of the scene in various bands.}
\label{fig9}
\vspace{-0.4cm} 
\setlength{\belowcaptionskip}{-0.5cm}
\end{figure}

\textbf{Spectral accuracy.}
We captured a spectral-interesting, texture-rich scene containing ColorCheck under D65 source illumination to evaluate the spectral accuracy of the hyperspectral image. The measurements captured by our prototype camera and the reconstruction result in the $495.4nm$ channel is shown in Figure \ref{fig9}(a). We also demonstrate excellent agreement between the reconstructed spectra at heart-shaped markers with intricate texture details and the corresponding ground truth spectra.

\textbf{Spatial resolution.}
In Figure \ref{fig9}(b), measurements of letters on the ColorCheck are compared with the reconstruction of the $495.4nm$ channel, which underwent reconstruction, markedly improving the MTF function.
Figure \ref{fig9}(c) demonstrates the successful reconstruction of the image within the yellow box, revealing clear textures in each band and restoring high-frequency details from aliased data.

\textbf{Tradeoff between accuracy and spectral resolution.} 
In ADIS, spectral resolution hinges on dispersion distance, while reconstruction accuracy is related to PSF concentration. A higher PSF dispersion decreases inter-band spectral data correlation, thereby alleviating underdetermination in the inverse process. Hence, future efforts should center on optimizing system parameters and algorithm performance to enhance overall performance.

\textbf{Sparse Propensity of Reconstruction.}
Comparing the reconstruction results of different scenes in Figure \ref{fig7}(b) and \ref{fig9}(c), the artifacts within ADIS reconstructions escalate when the texture complexity and spectral complexity intensify, 
which could potentially be mitigated through augmentation of training data complexity and diversity.

\textbf{}
\vspace{-0.3cm}
\section{Conclusion}
\vspace{-0.1cm}
A compact diffractive optical system comprising an ultra-thin aperture mask and conventional imaging lens forms a discrete coding pattern on a mosaic sensor. The Cascaded Shift-Shuffle Spectral Transformer (CSST) algorithm is used to decode the diffraction pattern for high-resolution hyperspectral imaging. Meanwhile, the system's spatial invariance ensures pattern robustness, and its diffraction efficiency is improved to 75\% using Babinet's principle. Further work is needed to improve imaging quality and spectral resolution while maintaining high diffraction efficiency. Furthermore, there's a need to investigate ADIS's potential for fulfilling large FOV demands.

\section{Acknowledgments}
\vspace{-0.1cm}
This work is supported by the National Natural  Science Foundation of China (No.62025108), the Leading Technology of Jiangsu Basic Research Plan (No.BK20192003), and the Key \& Plan of Jiangsu Province (No. BE2022155).


{\small
\bibliographystyle{unsrt}
\bibliography{egpaper_final}
}

\begin{figure*}[t]
  \includegraphics[width=1\linewidth]{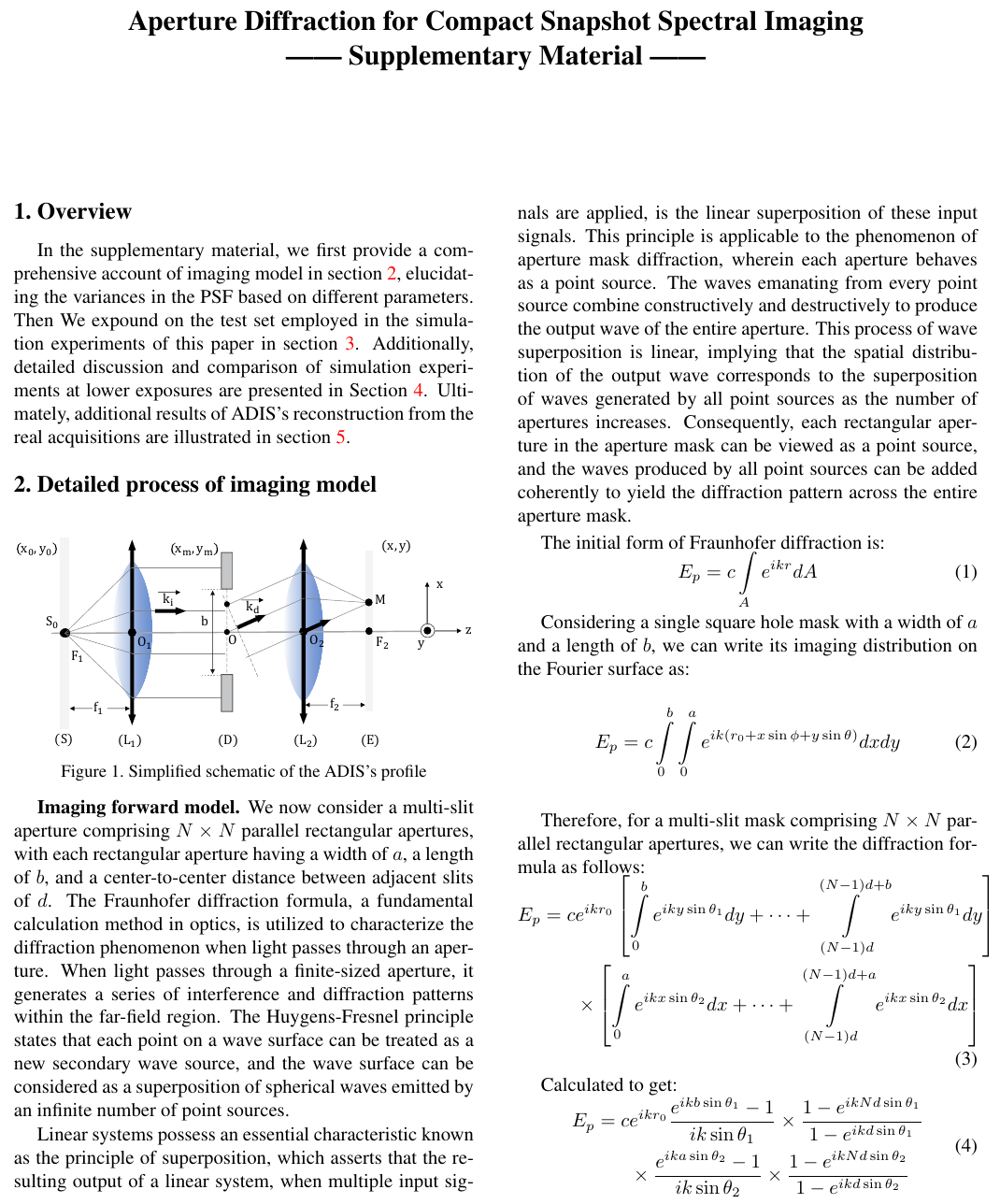}
\label{figex}
\end{figure*}

\begin{figure*}[t]
  \includegraphics[width=1\linewidth]{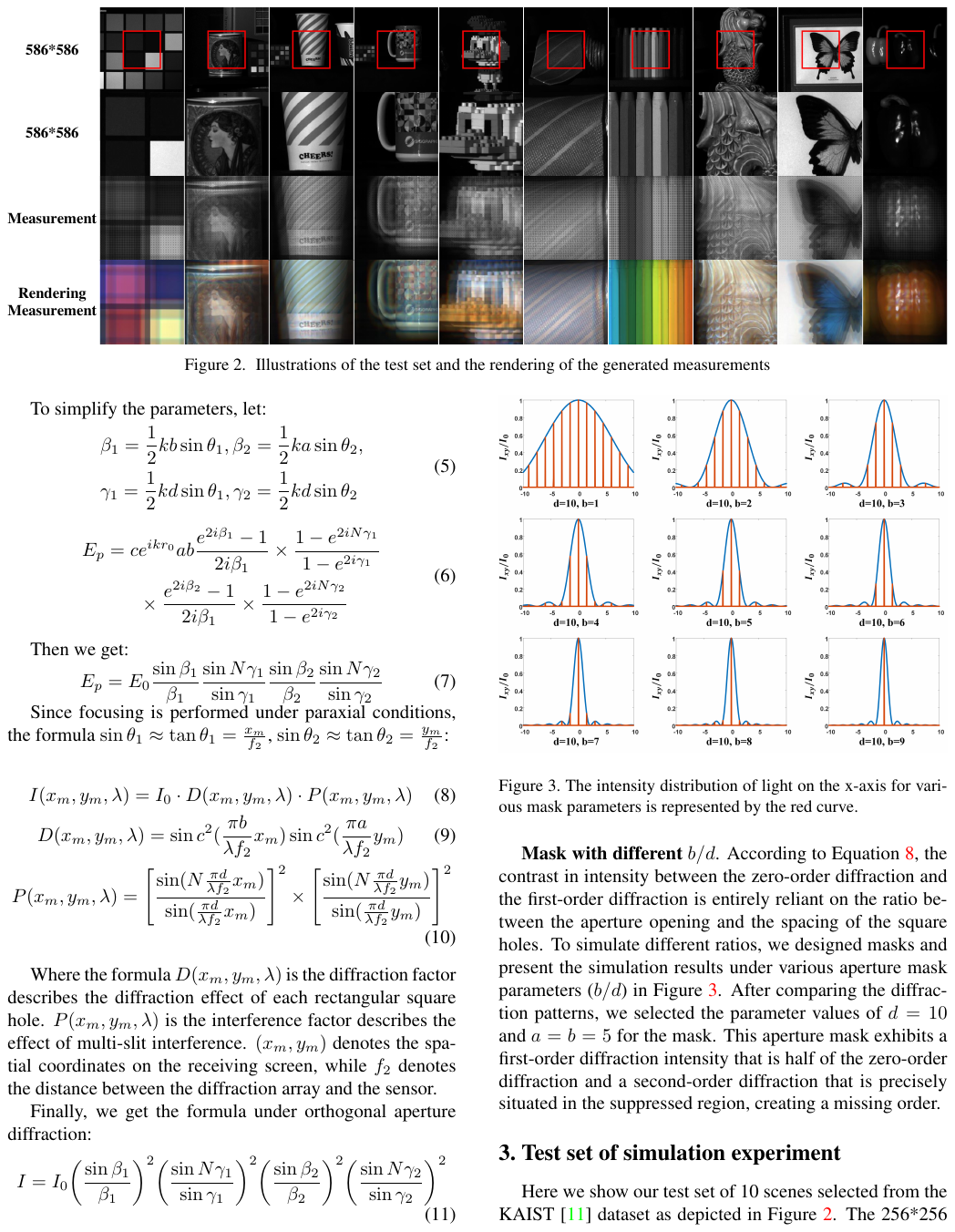}
\label{figex}
\end{figure*}

\begin{figure*}[t]
  \includegraphics[width=1\linewidth]{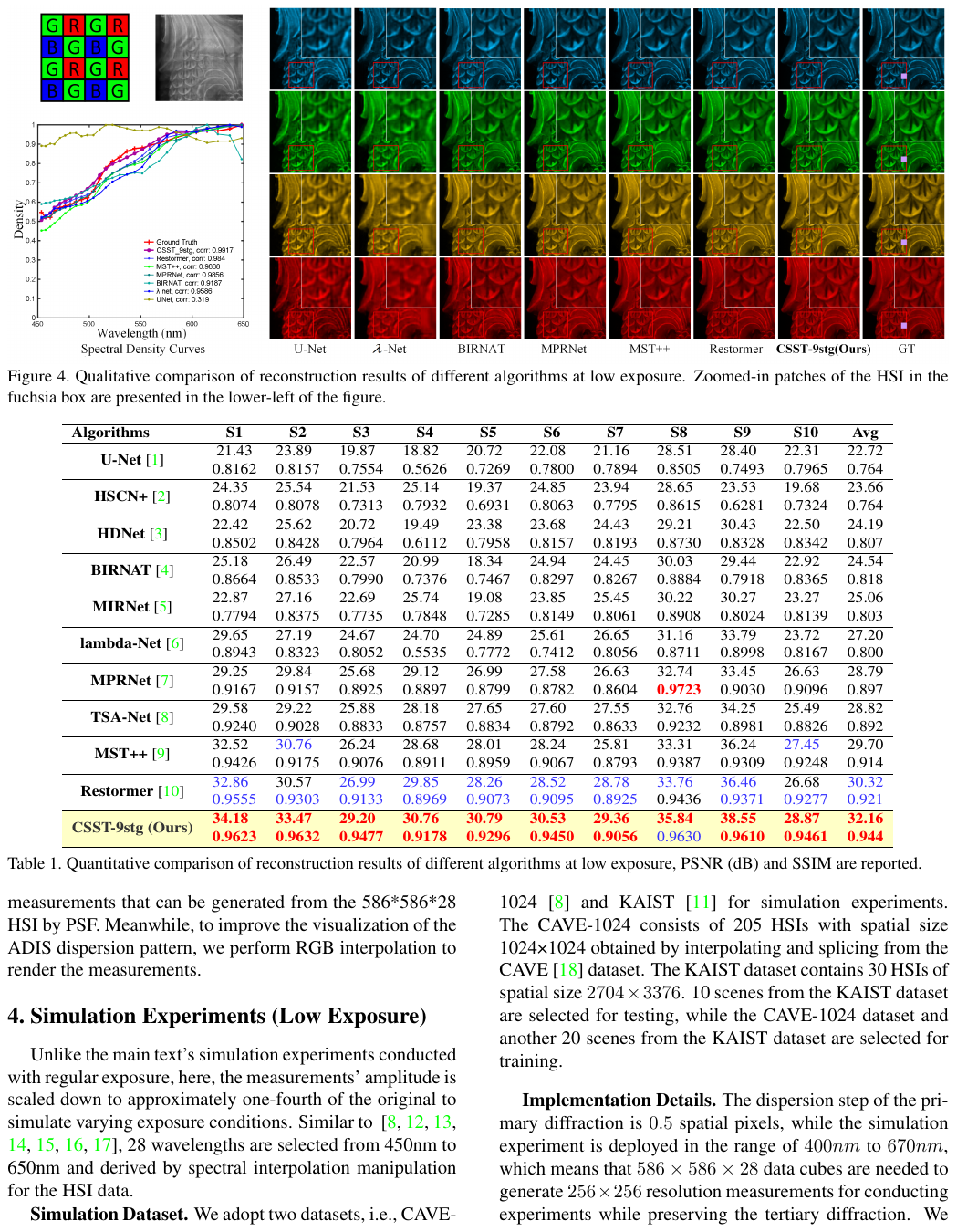}
\label{figex}
\end{figure*}

\begin{figure*}[t]
  \includegraphics[width=1\linewidth]{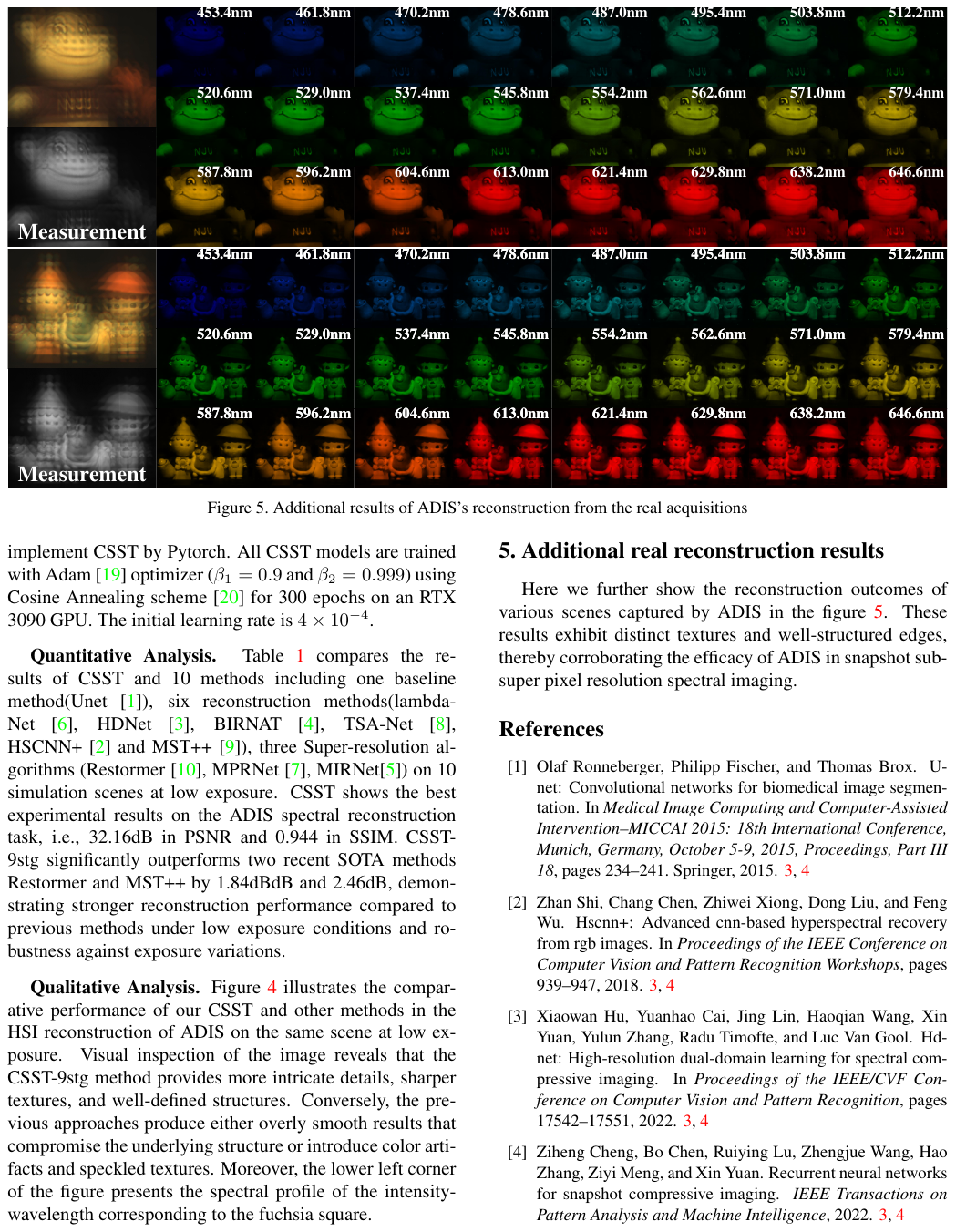}
\label{figex}
\end{figure*}

\begin{figure*}[t]
  \includegraphics[width=1\linewidth]{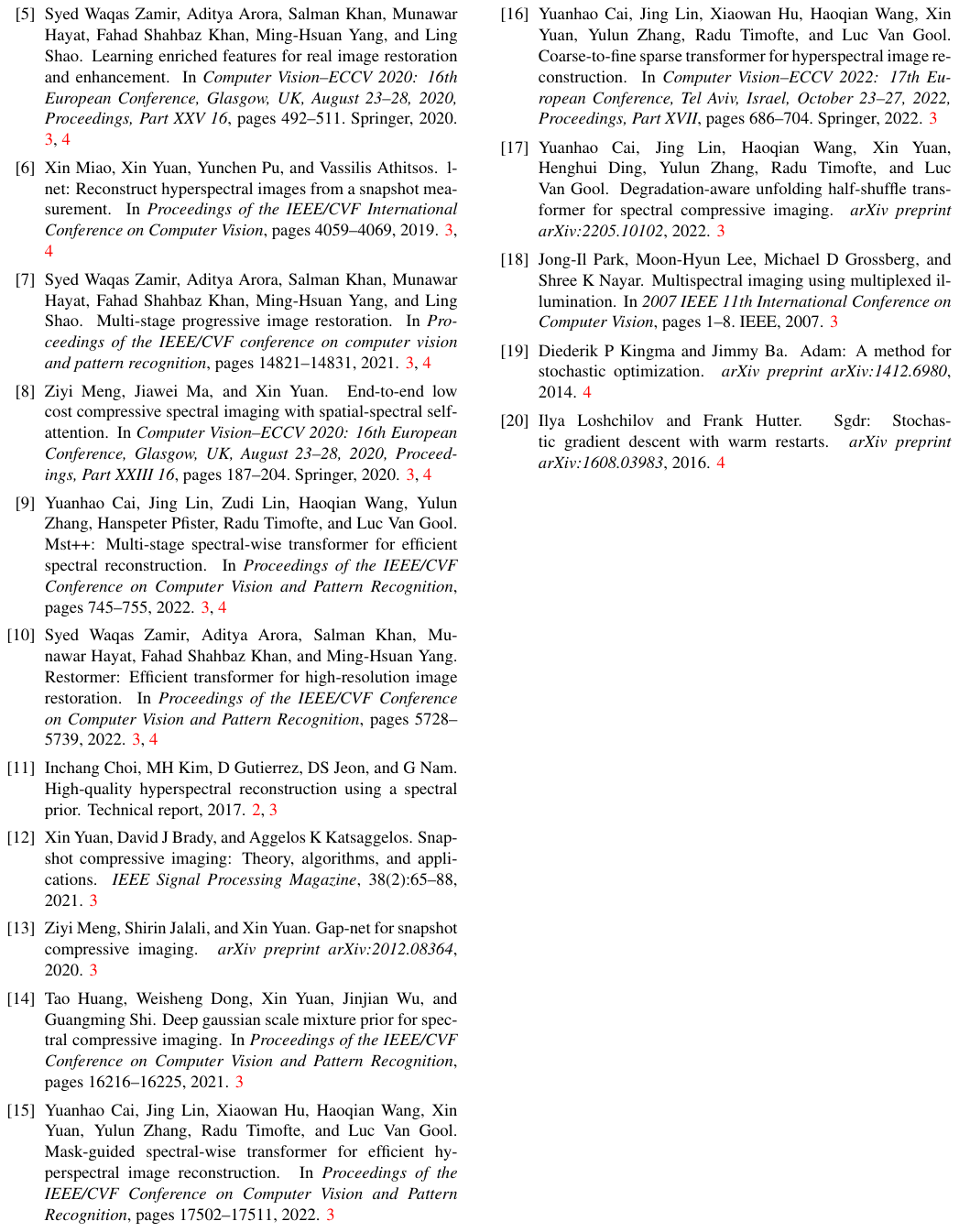}
\label{figex}
\end{figure*}

\end{document}